\documentclass[11pt,a4paper]{article}
\usepackage{times,latexsym}
\usepackage{url}

\usepackage[acceptedWithA]{tacl2018v2}

\usepackage{enumitem}
\usepackage{pifont}
\usepackage{textcomp}
\usepackage{subfigure}
\usepackage{amsmath,amssymb}

\usepackage{varwidth}
\usepackage{booktabs}
\usepackage{arydshln}
\usepackage{graphicx}
\usepackage{multirow}
\usepackage[T1]{fontenc}

\usepackage{bm}
\usepackage{xcolor}
\usepackage[font=footnotesize]{caption}

\usepackage{amssymb} 

\newif\ifcomments
\commentstrue 
\ifcomments
    \providecommand\at[1]{[\textcolor{blue}{{AT: #1}}]}
    \providecommand\jb[1]{[\textcolor{red}{{JB: #1}}]}
    \providecommand\ye[1]{[\textcolor{cyan}{{YE: #1}}]}
    \providecommand\yg[1]{[\textcolor{violet}{{YG: #1}}]}
\else
    \providecommand\at[1]{}
    \providecommand\jb[1]{}
    \providecommand\ye[1]{}
    \providecommand\yg[1]{}
\fi
\newcommand\comment[1]{}

\newcommand\nl[1]{\emph{``#1''}}
\newcommand\robertal{\textsc{RoBERTa-L}}

\newcommand\bertwwm{\textsc{BERT-WWM}}
\newcommand\bertl{\textsc{BERT-L}}

\newcommand\bert{\textsc{BERT}}

\newcommand\roberta{\textsc{RoBERTa}}

\newcommand\mlmbaseline{\textsc{MLM-Baseline}}

\newcommand\agecomp{\textsc{Age-Compare}}

\newcommand\nolang{\textsc{No Lang.}}

\newcommand\mcmlm{\textsc{MC-MLM}}
\newcommand\mcqa{\textsc{MC-QA}}

\newcommand\smetric{\textsc{WS}}
\newcommand\maxmetric{\textsc{Max}}

\newcommand\inputtokens{\bm{x}}
\newcommand\reprs{\bm{h}}
\newcommand\MLP{\text{MLP}_\text{MLM}}
\newcommand\MLPQA{\text{MLP}_\text{QA}}
\newcommand\linear{\textsc{Linear}}
\newcommand\nolanguage{\textsc{No Lang.}}
\newcommand\pertlanguage{\textsc{Perturbed Lang.}}
\newcommand\langsense{Language Sensitivity}

\newcommand\ansa{\textbf{A.}}
\newcommand\ansb{\textbf{B.}}
\newcommand\ansc{\textbf{C.}}
\newcommand\ansd{\textbf{D.}}
\newcommand\anse{\textbf{E.}}

\newcommand\nolangs{nolang}
\newcommand\pertlangs{pert}
\newcommand\langsenses{\textsc{LangSense}}
\newcommand\bfemph[1]{\textbf{\emph{#1}}}

\newcommand\semicheck{\checkmark\kern-1.1ex\raisebox{.7ex}{\rotatebox[origin=c]{125}{--}}}

\title{oLMpics - On what Language Model Pre-training Captures}

\author{
 Alon Talmor$^{1,2}$ ~~~~~ Yanai Elazar$^{1,3}$ ~~~~~
 Yoav Goldberg$^{1,3}$ ~~~~~
 Jonathan Berant$^{1,2}$ \\
 $^1$The Allen Institute for AI\\
$^2$Tel-Aviv University \\
$^3$Bar-Ilan University \\
  {\sf \{alontalmor@mail,joberant@cs\}.tau.ac.il} \\
  {\sf \{yanaiela,yoav.goldberg\}@gmail.com} \\
}

\begin{document}
\maketitle

\begin{abstract}

Recent success of pre-trained language models (LMs) has spurred widespread interest in the language capabilities that they possess. However, efforts to understand whether LM representations are useful for symbolic reasoning tasks have been limited and scattered. In this work, we propose eight reasoning tasks, which conceptually require operations such as comparison, conjunction, and composition. A fundamental challenge is to understand whether the performance of a LM on a task should be attributed to the pre-trained representations or to the process of fine-tuning on the task data. To address this, we propose an evaluation protocol that includes both zero-shot evaluation (no fine-tuning), as well as comparing the learning curve of a fine-tuned LM to the learning curve of multiple controls, which paints a rich picture of the LM capabilities. Our main findings are that: (a) different LMs exhibit qualitatively different reasoning abilities, e.g., \roberta{} succeeds in reasoning tasks where \bert{}  fails completely; (b) LMs do not reason in an abstract manner and are \emph{context-dependent}, e.g., while \roberta{} can compare ages, it can do so only when the ages are in the typical range of human ages; (c) On half of our reasoning tasks all models fail completely. Our findings and infrastructure can help future work on designing new datasets, models and objective functions for pre-training.

\comment{

\at{still a draft, please ingore this ... } 
The is a lot of interest in probing what knowledge is captured by LMs. current work mostly are based on measure on the ability to fine-tune a model. however fine tuning is not enough ... 

When trying to do this...


In this work, we introduce multiple methods and controls to []pin-point wheather specific types of reasonign are captured by the LMs, and to] disentangle knowledge gainerd by fine-tuning from interenal knowlege


We are parrticularly interested in the ability of LMs to capture various types of low level symbolic reasonings. To this end, we introduce seven tasks that require capabilities ranging from conjunction and composition to commonsense reasoning and comparison.

We find that the current pre-training objective does not provide the LM adequate tools to solve the probes and pass the controls. 

\robertal{} shows promising comparison capabilities, but fails to generalize to new values and proves to be sensitive to the context of the task at hand. 

Overall, the few reasoning capabilities that are observed, are strongly tied to language commonly found in the pre-training corpora. 

suggest our tasks as benchmark for future
}
\end{abstract}
\section{Introduction}

\begin{figure}[t]
  \includegraphics[width=\columnwidth]{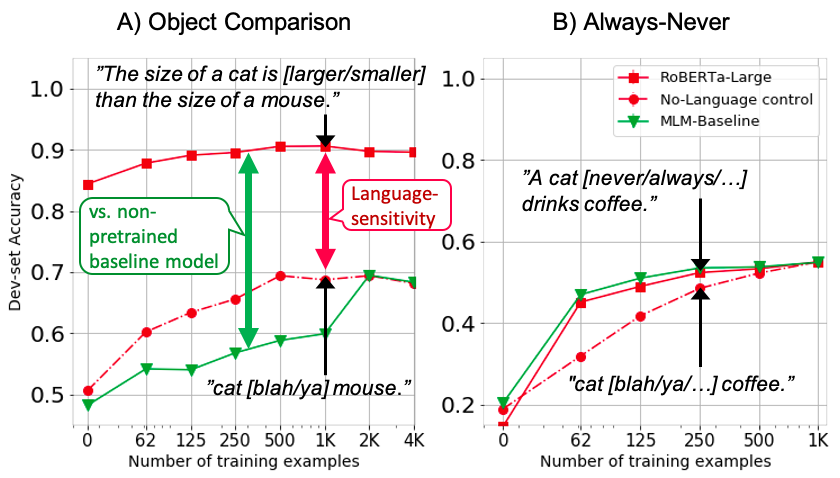}
  \caption{Overview of our experimental design. Two probes are evaluated using learning curves (including zero-shot). \robertal{}'s (red squares, upper text in black) accuracy is compared to a \textsc{No Language} (\nolang{}) control (red circles, lower text in black), and \mlmbaseline{}, which is not pre-trained (green triangles). Here, we conclude that the LM representations are well-suited for task A), whereas in task B) the model is adapting to the task during fine-tuning.}~\label{fig:intro-fig}
\end{figure}

Large pre-trained language models (LM) have revolutionized the field of natural language processing in the last few years \cite{NIPS2015_5949,peters2018elmo,yang2019xlnet,radford2019gpt2,devlin2019bert}.
This has instigated research exploring what is captured by the contextualized representations that these LMs compute, revealing that they encode substantial amounts of syntax and semantics \cite{linzen2016assessing,tenney2018what,tenney-etal-2019-bert,lexcomp_tacl_2019,lin2019open,coenen2019visualizing}.

\begin{table*}[t]
\centering
\resizebox{1.0\textwidth}{!}{
\begin{tabular}{l|l|l|c}
 \textbf{Probe name} & \textbf{Setup} & \textbf{Example} & \textbf{Human\footnotemark[1]} \\ 
\hline
\textsc{Always-Never} & \mcmlm{} &  \emph{A \underline{chicken} [MASK] has \underline{horns}.} \ansa{} never \ansb{} rarely \ansc{} sometimes \ansd{} often \anse{} always  & 91\% \\
\hdashline 
\textsc{Age Comparison} & \mcmlm{} & \emph{A \underline{21} year old person is  [MASK] than me in age, If I am a \underline{35} year old person.} \ansa{} younger \ansb{} older  & 100\% \\
\textsc{Objects Comparison} & \mcmlm{} & \emph{The size of a \underline{airplane} is [MASK] than the size of a \underline{house} .} \ansa{} larger \ansb{} smaller & 100\% \\ 
\hdashline
\textsc{Antonym Negation} & \mcmlm{} & \emph{It was [MASK] \underline{hot}, it was really \underline{cold} .} \ansa{} not \ansb{} really  & 90\% \\
\hdashline

\textsc{Property Conjunction} & \mcqa{} & \emph{What is usually \underline{located at hand} and \underline{used for writing}?} \ansa{} pen \ansb{} spoon \ansc{} computer & 92\% \\
\textsc{Taxonomy Conjunction} & \mcmlm{} &  \emph{A \underline{ferry} and a \underline{floatplane} are both a type of [MASK].} \ansa{} vehicle \ansb{} airplane \ansc{} boat & 85\% \\
\hdashline
\textsc{Encyc. Composition} & \mcqa{} & \emph{When did the band where Junior Cony played first form?} \ansa{} 1978 \ansb{} 1977 \ansc{} 1980 & 85\% \\
\textsc{Multi-hop Composition} & \mcmlm{} & \emph{When comparing a \underline{23}, a \underline{38} and a \underline{31} year old, the [MASK] is oldest} \ansa{} second \ansb{} first \ansc{} third & 100\% \\
\end{tabular}}
\caption{Examples for our reasoning probes. We use two types of experimental setups, explained in \S\ref{sec:models}.  A. is the correct answer. }
\label{tab:intro}
\end{table*}

Despite these efforts, it remains unclear \emph{what symbolic reasoning capabilities are difficult to learn from an LM objective only?} In this paper, we propose a diverse set of probing tasks for types of symbolic reasoning that are potentially difficult to capture using a LM objective (see Table~\ref{tab:intro}). Our intuition is that since a LM objective focuses on word co-occurrence, it will struggle with tasks that are considered to involve symbolic reasoning such as determining whether a \emph{conjunction} of properties is held by an object, and \emph{comparing} the sizes of different objects. Understanding what is missing from current LMs may help design datasets and objectives that will endow models with the missing capabilities.

However, how does one verify whether pre-trained representations hold information that is useful for a particular task?
Past work mostly resorted to fixing the representations and \emph{fine-tuning} a simple, often linear, randomly initialized probe, to determine whether the representations hold relevant information \cite{ettinger2016probing,adi2016fine,belinkov2019analysis,structural-probe,wallace2019nlp,rozen2019diversify,peters2018dissecting,warstadt2019investigating}. However, it is difficult to determine whether success is due to the pre-trained representations or due to fine-tuning itself \cite{hewitt2019designing}.
To handle this challenge, we include multiple controls that improve our understanding of the 
results.

Our ``purest'' setup is zero-shot: we cast tasks in the \emph{masked LM} format, and use a pre-trained LM without any fine-tuning. For example, given the statement \nl{A cat is [MASK] than a mouse}, an LM can decide if the probability of \nl{larger} is higher than \nl{smaller} for the a masked word (Figure~\ref{fig:intro-fig}). If a model succeeds without pre-training over many pairs of objects, then its representations are useful for this task. However, if it fails, it could be due to a mismatch between the language it was pre-trained on and the language of the probing task (which might be automatically generated, containing grammatical errors). Thus, we also compute the learning curve (Figure~\ref{fig:intro-fig}), 
by fine-tuning with increasing amounts of data on the already pre-trained masked language modeling (MLM) output ``head'', a 1-hidden layer MLP on top of the model's contextualized representations.  A model that adapts from fewer examples arguably has better representations for it. 

Moreover, to diagnose whether model performance is related to pre-training or fine-tuning, we add controls to every experiment (Figures~\ref{fig:intro-fig},\ref{fig:controls-example}). First, we add a control 
that makes minimal use 
of language tokens, i.e., \nl{cat [MASK] mouse} (\nolang{} in  Figure~\ref{fig:intro-fig}).  
If a model succeeds given minimal use of language, the
performance can be mostly attributed to fine-tuning rather than to the pre-trained language representations. Similar logic is used to compare against baselines that are not pre-trained (except for non-contextualize word embeddings).
Overall, our setup provides a rich picture of whether LM representations help in solving a wide range of tasks.

We introduce eight tasks that test different types of reasoning, as shown in Table~\ref{tab:intro}.\footnote{Average human accuracy was evaluated by two of the authors. Overall inter-annotator agreement accuracy was 92\%.}
We run experiments using several pre-trained LMs, based on \bert{} \cite{devlin2019bert} and \roberta{} \cite{liu2019roberta}. We find that there are clear qualitative differences between different LMs with similar architecture. For example, \textsc{RoBERTa-Large} (\robertal{}) can perfectly solve some reasoning tasks, such as comparing numbers, even in a zero-shot setup, while other models' performance is close to random. However, good performance is highly \emph{context-dependent}. Specifically, we repeatedly observe that even when a model solves a task, small changes to the input quickly derail it to low performance. For example, \robertal{} can almost perfectly compare people's ages, when the numeric values are in the expected range (15-105), but miserably fails if the values are outside this range. 
Interestingly, it is able to reliably answer when ages are specified through the birth year in the range 1920-2000.
This highlights that the LMs ability to solve this task is strongly tied to the specific values and linguistic context and does not generalize to arbitrary scenarios. 
Last, we find that in four out of eight tasks, all LMs perform poorly compared to the controls. 

 

Our contributions are summarized as follows:
\begin{itemize}[leftmargin=*,topsep=0pt,itemsep=0pt,parsep=0pt]
    \item A set of probes that test whether specific reasoning skills are 
    captured by pre-trained LMs.
    \item An evaluation protocol for understanding whether a capability is encoded in pre-trained representations or is learned during fine-tuning.
    \item An analysis of skills that current LMs possess. We find that LMs with similar architectures are qualitatively different, that their success is context-dependent, and that often all LMs fail.
    \item Code and infrastructure for designing and testing new probes on a large set of pre-trained LMs.
\end{itemize}

The code and models are available at \url{http://github.com/alontalmor/oLMpics}.

\section{Models}
\label{sec:models}
We now turn to the architectures and loss functions used throughout the different probing tasks.

\subsection{Pre-trained Language Models}

All models in this paper take a sequence of tokens $\inputtokens = (\inputtokens_1, \dots, \inputtokens_n)$, and compute contextualized representations with a pre-trained LM, that is, $\reprs = \text{ENCODE}(\inputtokens) = (\reprs_1, \dots, \reprs_n)$. 
Specifically, we consider (a) \bert{}: \cite{devlin2019bert}, a pre-trained LM built using the Transformer \cite{vaswani2017attention} architecture, which consists of a stack of Transformer layers, where each layer includes a multi-head attention sublayer and a feed-forward sub-layer. BERT is trained on large corpora using the \emph{masked-language modeling} objective (MLM), i.e., the model is trained to predict words that are masked from the input; including \textsc{BERT-Whole-Word-Masking} (\bertwwm{}), that was trained using \emph{whole-word-masking} (b) \textsc{RoBERTa} \cite{liu2019roberta}, which has the same architecture as \bert{}, but was trained on 10x more data and optimized carefully.





\subsection{Probing setups}
We probe the pre-trained LMs using two setups:
multi-choice masked LM (\mcmlm{}) and multi-choice question answering (\mcqa{}). The default setup is \mcmlm{}, used for tasks where the answer-set is small, consistent across the different questions, and each answer appears as a single item in the word-piece vocabulary.\footnote{Vocabularies of LMs such as \bert{} and \roberta{} contain \emph{word-pieces}, which are sub-word units that are frequent in the training corpus. For details see \newcite{sennrich2015neural}.} The MC-QA setup is used when the answer-set substantially varies between questions, and many of the answers have more than one word piece.

\paragraph{\mcmlm{}}
Here, we convert the MLM setup to a multi-choice setup (\mcmlm{}). Specifically, the input to the LM is the sequence $\inputtokens = (\texttt{[CLS]}, \dots, \inputtokens{}_{i-1}, \texttt{[MASK]}, \inputtokens{}_{i+1}, \dots, \texttt{[SEP]})$, where a single token $\inputtokens_i$ is masked. Then, the contextualized representation $\reprs_i$ is passed through a \emph{\mcmlm{} head} where $\mathcal{V}$ is the vocabulary, and $FF_\text{MLM}$ is a 1-hidden layer MLP:
$$
l = FF_\text{MLM}(\reprs_i) \in \mathbb{R}^{|\mathcal{V}|},
p = \text{softmax}(m \oplus l),
$$
where $\oplus$  is element-wise addition and $m \in \{0, -\infty\}^{|\mathcal{V}|}$ is a mask that
guarantees that the support of the probability distribution will be over exactly $K \in \{2,3,4,5\}$ candidate tokens: the correct one and $K-1$ distractors. Training minimizes cross-entropy loss given the gold masked token.
An input, e.g. \nl{\texttt{[CLS]} Cats \texttt{[MASK]} drink coffee \texttt{[SEP]}}, is passed through the model, the contextualized representation of the masked token is passed through the MC-MLM head, and the final distribution is over the vocabulary words \nl{always}, \nl{sometimes} and \nl{never}, where the gold token is \nl{never}, in this case. 

A compelling advantage of this setup, 
is that reasonable performance can be obtained without training, using the original LM representations and the already pre-trained MLM head weights \cite{petroni2019language}.


\paragraph{\mcqa{}}
Constructing a MC-MLM probe limits the answer candidates to a single token from the word-piece vocabulary. To relax this we use in two tasks the standard setup for answering multi-choice questions with pre-trained LMs \cite{talmor2019commonsenseqa, OpenBookQA2018}.
Given a question $\bm{q}$ and candidate answers $\bm{a}_1, \dots, \bm{a}_K$, we compute for each candidate answer $\bm{a}_k$ representations $\bm{h}^{(k)}$ from the input tokens \nl{\texttt{[CLS]} $\bm{q}$ \texttt{[SEP]} $\bm{a}_k$ \texttt{[SEP]}}. Then the probability over answers is obtained using the \emph{multi-choice QA head}:
$$
l^{(k)} = FF_\text{QA}(\reprs_1^{(k)}),
p = \text{softmax}(l^{(1)}, \dots, l^{(K)}),
$$
where $FF_\text{QA}$ is a 1-hidden layer MLP that is run over the \texttt{[CLS]} (first) token of an answer candidate and outputs a single logit. Note that in this setup that parameters of $FF_\text{QA}$ cannot be initialized using the original pre-trained LM.

\subsection{Baseline Models}
To provide a lower bound on the performance of pre-trained LMs, we introduce two baseline models with only non-contextualized representations. 


\paragraph{\mlmbaseline{}}
This serves as a lower-bound for the \mcmlm{} setup.
The input to $FF_\text{MLM}(\cdot)$ is the hidden representation $\reprs \in \mathbb{R}^{1024}$ (for large models). To obtain a similar architecture with non-contextualized representations, we concatenate the first $20$ tokens of each example, representing each token with a $50$-dimensional \textsc{GloVe} vector \cite{pennington2014glove}, and pass this $1000$-dimensional representation of the input through $FF_\text{MLM}$, exactly like in \mcmlm{}. In all probes, phrases are limited to 20 tokens. If there are less than 20 tokens in the input, we zero-pad the input.

\paragraph{\mcqa{} baseline}
This serves as a lower-bound for \mcqa{}.
We use the ESIM architecture over \textsc{GloVe}
representations, which is known to provide a strong model when the input is a pair of text fragments \cite{chen2017enhanced}. We adapt the architecture to the multi-choice setup using the procedure proposed by \newcite{zellers2018swag}. Each phrase and candidate answer are passed as a list of token \texttt{`[CLS] phrase [SEP] answer [SEP]'} to the LM. The contextualized representation of the \texttt{[CLS]} token is linearly projected to a single logit. The logits for candidate answers are passed through a softmax layer to obtain  probabilities, and the argmax is selected as the model prediction.


\section{Controlled Experiments}
\label{sec:controlled_experiments}
We now describe the experimental design and controls used to interpret the results.
We use the \agecomp{} task as a running example, where models need to compare the numeric value of ages. 
\subsection{Zero-shot Experiments with \mcmlm{}}
\label{subsec:zero_shot}
Fine-tuning pre-trained LMs makes it hard to disentangle what is captured by the original representations and what was learned during fine-tuning. Thus, ideally, one should test LMs using the pre-trained weights \emph{without}
fine-tuning \cite{linzen2016agreement,goldberg2019assessing}.
The \mcmlm{} setup, which uses a pre-trained MLM head, achieves exactly that. One only needs to design the task as a statement with a single masked token and $K$ possible output tokens. For example, in \agecomp{}, we chose the phrasing \nl{A \texttt{AGE-1} year old person is  \texttt{[MASK]} than me in age, If I am a \texttt{AGE-2} year old person.}, where \texttt{AGE-1} and \texttt{AGE-2} are replaced with different integers, and 
possible answers are \nl{younger} and \nl{older}. Otherwise, no training is needed, and the original representations are tested.

Figure~\ref{fig:controls-example}A provides an example of such zero-shot evaluation.  Different values are assigned to \texttt{AGE-1} and \texttt{AGE-2}, and the pixel is colored when the model predicts \nl{younger}. Accuracy (acc.) is measured as the proportion of cases when the model output is correct.
The performance of \bertwwm{}, is on the left (blue), and of \robertal{} on the right (green). The results in Figure~\ref{fig:controls-example}A and Table~\ref{tab:agecomparison} show that \robertal{} compares numbers correctly (98\% acc.), \bertwwm{} achieves higher than random acc. (70\% acc.), while \bertl{} is random (50\% acc.). 
The performance of \mlmbaseline{} is also random,
as the $\MLP{}$ weights are randomly initialized.

We note that picking the statement for each task was done through manual experimentation. We tried multiple phrasings \cite{jiangHowCanWe2019} and chose the one that achieves highest average zero-shot accuracy across all tested LMs.

A case in point ... 

Thus, if a model performs well, one can infer that it has the tested reasoning skill. However, failure does not entail that the reasoning skill is missing, as it is possible that there is a problem with the lexical-syntactic construction we picked.  



\subsection{Learning Curves}
Despite the advantages of zero-shot evaluation, performance of a model might be adversely affected by mismatches between the language the pre-trained LM was trained on and the language of the examples in our tasks \cite{jiangHowCanWe2019}. 

To tackle this, we fine-tune models  with a small number of examples.
We assume that if the LM representations are useful for a task, it will require few examples to overcome the language mismatch and achieve high performance. In most
cases, we train with $N \in \{62, 125, 250, 500, 1K, 2K, 4K\}$ examples. 
To account for optimization instabilities, we fine-tune several times with different seeds,
and report average accuracy across seeds.
The representations $\reprs{}$ are fixed during fine-tuning, and we only fine-tune the parameters of $\MLP{}$.



\begin{table}[t]
\centering
\resizebox{1.0\columnwidth}{!}{
\begin{tabular}{l|c|cc|cc|cc}
 Model & Zero & \multicolumn{2}{c|}{$\MLP{}$} & \multicolumn{2}{c|}{\linear{}}
 & \multicolumn{2}{c}{\langsenses{}}  \\ 
\toprule
& shot &\smetric{}&\maxmetric{}&\smetric{}&\maxmetric{}&\pertlangs{}& \nolangs{}\\
\midrule
RoBERTa-L &  98 &  98 &  100 &  97 &  100 &  31 &  51 \\
BERT-WWM  &  70 &  82 &  100 &  69 &   85 &  13 &  15 \\
BERT-L    &  50 &  52 &   57 &  50 &   51 &   1 &   0 \\
\hdashline
RoBERTa-B &  68 &  75 &   91 &  69 &   84 &  24 &  25 \\
BERT-B    &  49 &  49 &   50 &  50 &   50 &   0 &   0 \\
\hdashline
Baseline  &  49 &  58 &   79 &   - &   - &   0 &   0 \\
\end{tabular}}
\caption{\agecomp{} results. Accuracy over two answer candidates (random is 50\%). \langsenses{} are the \langsense{} controls, \pertlangs{} is \pertlanguage{} and \nolangs{} is \nolanguage{}. The baseline row is \mlmbaseline{}. } 
\label{tab:agecomparison}
\end{table}

\paragraph{Evaluation and learning-curve metrics}
Learning curves are informative, but inspecting many learning curves can be difficult. Thus, we summarize them using two aggregate statistics.
We report: (a) \textsc{Max}, i.e., the maximal accuracy on the learning curve, used to estimate how well the model can handle the task given the limited amount of examples. (b) The metric \smetric{}, which is a weighted average of accuracies across the learning curve, where higher weights are given to points where $N$ is small.\footnote{ 
We use the decreasing weights 
$W=(0.23, 0.2, 0.17, 0.14, 0.11, 0.08, 0.07)$. } 
\smetric{} is related to the area under the accuracy curve, and to the online code metric, proposed by \citet{yogatama2019learning,blier2018description}. The linearly decreasing weights emphasizes our focus on performance given little training data, as it highlights what was encoded by the model \emph{before} fine-tuning.

For \agecomp{}, the solid lines in Figure~\ref{fig:controls-example}B illustrate the learning curves of \robertal{} and \bertwwm{}, and Table~\ref{tab:agecomparison} shows the aggregate statistics. We fine-tune the model by replacing \texttt{AGE-1} and \texttt{AGE-2} with values between $43$ and $120$, but test with values between $15$ and $38$, to guarantee that the model \emph{generalizes} to values unseen at training time. Again, we see that the representations learned by \robertal{} are already equipped with the knowledge necessary for solving this task.   

%


\subsection{Controls}

Comparing learning curves tells us which model learns from fewer examples. 
However, since highly-parameterized MLPs, as used in LMs, can approximate a wide range of functions, it is difficult to determine whether performance is tied to the knowledge acquired at pre-training time, or to the process of fine-tuning itself. We present controls that attempt to disentangle these two factors. 

\begin{figure}[t]
  \includegraphics[width=\columnwidth]{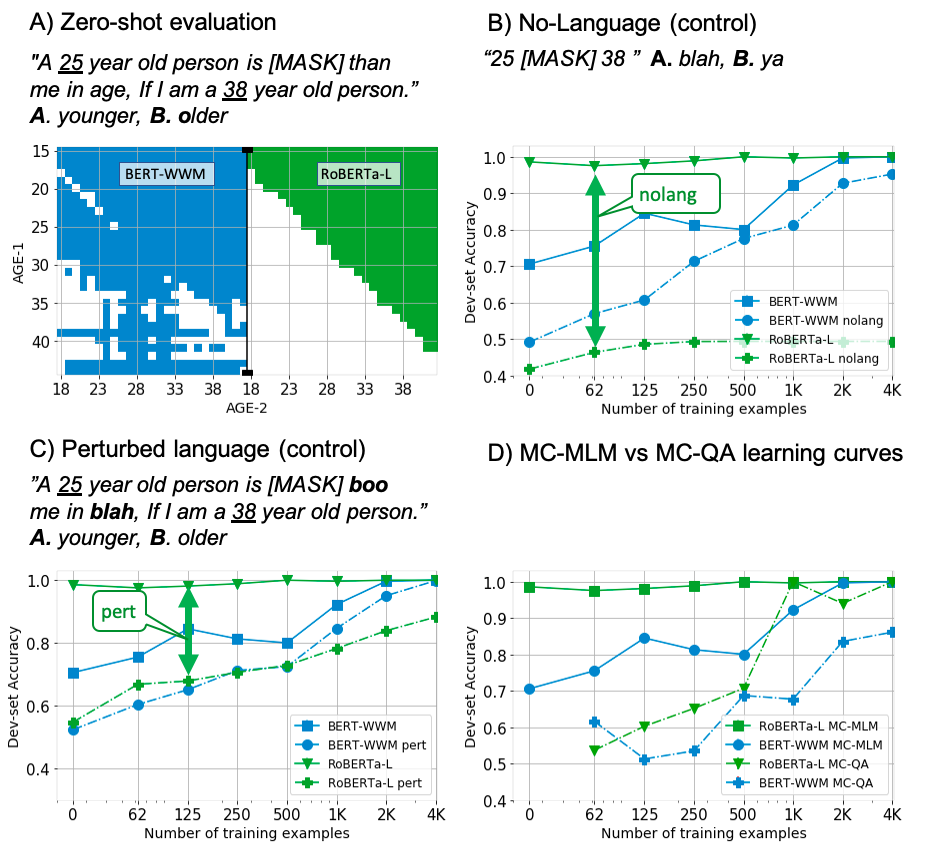}
  \caption{An illustration of our evaluation protocol. We compare \robertal{} (green) and \bertwwm{} (blue), controls are in dashed lines and markers are described in the legends. Zero-shot evaluation on the top left, \texttt{AGE-1} is \nl{younger} (in color) vs. \nl{older} (in white) than \texttt{AGE-2}.}~\label{fig:controls-example}
\end{figure}

\paragraph{Are LMs sensitive to the language input?}
We are interested in whether pre-trained representations reason over language examples. Thus, a natural control is to present the reasoning task \emph{without} language and inspect performance. If the learning curve of a model does not change when the input is perturbed or even mostly deleted, then the model shows low \emph{language sensitivity} and the pre-trained representations do not explain the probe performance.
This approach is related to work by \newcite{hewitt2019designing}, who proposed a control task, where the learning curve of a model is compared to a learning curve when words are associated with random behaviour. We propose two control tasks:

\noindent
\textbf{\emph{\textsc{No Language} control}}
We remove all input tokens, except for \texttt{[MASK]} and the \emph{arguments} of the task, i.e., the tokens that are necessary for computing the output. 
In \agecomp{}, an example is reduced to the phrase \nl{24 [MASK] 55}, where the candidate answers are the words \nl{blah}, for \nl{older}, and \nl{ya}, for \nl{younger}.
If the learning curve is similar to when the full example is given (low language sensitivity), then the LM is not strongly using the language input.

The dashed lines in Figure~\ref{fig:controls-example}B illustrate the learning curves in \nolanguage{}: \robertal{} (green) shows high language sensitivity, while \bertwwm{} (blue) has lower language sensitivity. This suggests it handles this task partially during fine-tuning. Table~\ref{tab:agecomparison} paints a similar picture, where the metric we use is identical to \smetric{}, except that instead of averaging accuracies, we average the \emph{difference} in accuracies between the standard model and \nolanguage{} (rounding negative numbers to zero). For \robertal{} the value is $51$, because \robertal{} gets almost $100\%$ acc. in the presence of language, and is random ($50\%$ acc.) without language. 




\noindent
\textbf{\emph{\textsc{Perturbed Language} control}} 
A more targeted language control, is to replace words that are central for the reasoning task with nonsense words. Specifically, we pick key words in each probe template, and replace these words by randomly sampling from a list of 10 words that carry relatively limited meaning.\footnote{The list of substitutions is: \nl{blah}, \nl{ya}, \nl{foo}, \nl{snap}, \nl{woo}, \nl{boo}, \nl{da}, \nl{wee}, \nl{foe} and \nl{fee}.}
For example, in \textsc{Property Conjunction}, we can replace the word \nl{and} with the word \nl{blah} to get the example \nl{What is located at hand \textbf{blah} used for writing?}. 
If the learning curve of \pertlanguage{} is similar to the original example, then the model does not utilize the pre-trained representation of \nl{and} to solve the task, and may not capture its effect on the semantics of the statement.

Targeted words change from probe to probe.
For example, in \agecomp{}, the targeted words are \nl{age} and \nl{than}, resulting in examples like \nl{A \texttt{AGE-1} year old person is [MASK] \textbf{blah} me in \textbf{da}, If i am a \texttt{AGE-2} year old person.}.
Figure~\ref{fig:controls-example}C shows the learning curves for \robertal{} and \bertwwm{}, where solid lines corresponds to the original examples and dashed lines are the \pertlanguage{} control.
Despite this minor perturbation, the performance of \robertal{} substantially decreases, implying that the model needs the input. Conversely, \bertwwm{} performance decreases only moderately.


\paragraph{Does a linear transformation suffice?}
In MC-MLM, the representations $\reprs{}$ are fixed, and only the pre-trained parameters of $\MLP{}$ are fine-tuned. As a proxy for measuring ``how far" the representations are from solving a task, we fix the weights of the first layer of $\MLP{}$, and only train the final layer. Succeeding in this setup means that only a linear transformation of $\reprs{}$ is required.
Table~\ref{tab:agecomparison} shows the performance of this setup (\linear{}), compared to $\MLP{}$. 
\paragraph{Why is \mcmlm{} preferred over \mcqa{}?} 
Constructing a MC-MLM probe limits the answer candidates to a single token from the word-piece vocabulary. To relax this setup we also explore the \mcqa{} setup from  \S\ref{sec:models}.
In MC-QA, we phrase the task as a question, letting answer candidates be arbitrary strings, which provides ample expressivity \cite{gardner2019question} and facilitates probing question involving complex and commonsense reasoning \cite{talmor2019commonsenseqa,gardner2019making,talmor2018web}. 
In Table~\ref{tab:intro},  \textsc{Property conjunction} and \textsc{Encyc. Comparison} serve as examples for this setup. For \agecomp{} we use the same task in \mcqa{} setup.
Figure~\ref{fig:controls-example}D compares the learning curves of \mcmlm{} and \mcqa{} in \agecomp{}.
Because in \mcqa{}, the network $\MLPQA{}$ cannot be initialized by pre-trained weights, zero-shot evaluation is not meaningful, and more training examples are needed to train $\MLPQA{}$. 
Still, the trends observed in \mcmlm{} remain, with \robertal{} achieving best performance with the fewest examples.

\section{The oLMpic Games}


We now move to describe the research questions and various probes used to answer these questions. For each task we describe how it was constructed, show results via a table as described in the controls section, and present an analysis.

Our probes are mostly targeted towards symbolic reasoning skills (Table~\ref{tab:intro}). We examine the ability of language models to compare numbers, to understand whether an object has a conjunction of properties, to perform multi-hop composition of facts, among others. However, since we generate examples automatically from existing resources, some probes also require background knowledge, such as sizes of objects. Moreover, as explained in \S\ref{subsec:zero_shot}, we test models on a manually-picked phrasing that might interact with the language abilities of the model. Thus, when a model succeeds this is evidence that it has the necessary skill, but failure could be attributed to issues with background knowledge and linguistic abilities as well. In each probe, we will explicitly mention what knowledge and language abilities are necessary.


\subsection{Can LMs perform robust comparison?}
\label{sec:num-comparison}
Comparing two numeric values requires representing the values and performing the comparison operations. In \S\ref{sec:controlled_experiments} we saw the \agecomp{} task, in which ages of two people were compared. We found that \robertal{} and to some extent \bertwwm{} were able to handle this task,  performing well under the controls.
We expand on this to related comparison tasks and perturbations that assess the sensitivity of LMs to the particular context and to the numerical value.

\begin{figure}[h]
  \includegraphics[width=\columnwidth]{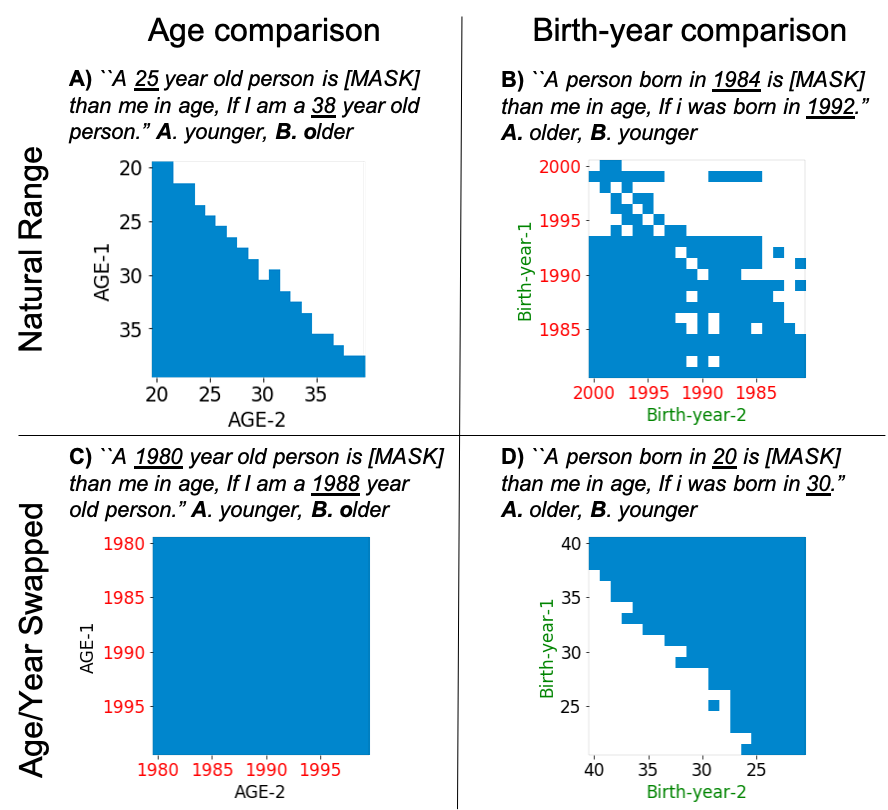}
  \caption{\textsc{Age comparison} perturbations. Left side graphs are age-comparison, right side graphs are age comparison by birth-year. In the bottom row, the values of ages are swapped with birth-years and vice versa. In blue pixels the model predicts \nl{older}, in white
  \nl{younger}. (A) is the correct answer.}~\label{fig:age-comp-pert}
\end{figure}

\paragraph{Is \robertal{} comparing numbers or ages?}
\robertal{} obtained zero-shot acc. of  98\% in \agecomp{}. But is it robust? We test this using perturbations to the task and present the results in  Figure~\ref{fig:age-comp-pert}. Figure~\ref{fig:age-comp-pert}A corresponds to the experiment from~\S\ref{sec:controlled_experiments},
where we observed that \robertal{} predicts \nl{younger} (blue pixels) and \nl{older} (white pixels) almost perfectly.

To test whether \robertal{} can compare ages given the birth year rather than the age, we use the statement \nl{A person born in \texttt{YEAR-1} is [MASK] than me in age, If i was born in \texttt{YEAR-2}.}   Figure~\ref{fig:age-comp-pert}B shows that 
it correctly flips \nl{younger} to \nl{older} (76\% acc.), reasoning that a person born in 1980 is older than one born in 2000.

However, when evaluated on the exact same statement, but with values corresponding to typical \emph{ages} instead of years (Figure~\ref{fig:age-comp-pert}D), \robertal{} obtains an acc. of 12\%, consistently outputting the opposite prediction.
With ages as values and not years, it seems to disregard the language, performing the comparison based on the values only. 
We will revisit this tendency in \S\ref{subsec:conjunction}.

Symmetrically, Figure~\ref{fig:age-comp-pert}C shows results when numeric values of ages are swapped with typical years of birth.
\robertal{} is unable to handle this, always predicting \nl{older}.\footnote{We observed that in neutral contexts models have a slight preference for \nl{older} over \nl{younger}, which could potentially explain this result.}
This emphasizes that the model is sensitive to the argument values.


\paragraph{Can Language Models compare object sizes?}
Comparing physical properties of objects requires knowledge of the numeric value of the  property and the ability to perform comparison. Previous work has shown that such knowledge can be extracted from text and images \cite{bagherinezhad2016elephants,forbes2017verb,yang2018extracting,elazar-etal-2019-large,pezzelle2019red}. Can LMs do the same?
\noindent
\bfemph{Probe Construction}
We construct statements of the form \nl{The size of a \texttt{OBJ-1} is usually much [MASK] than the size of a \texttt{OBJ-2}.}, where the candidate answers are \nl{larger} and \nl{smaller}.
To instantiate the two objects, we manually sample from a list of objects from two domains: animals (e.g. \nl{camel}
) and general objects (e.g. 
\nl{sun}), and use the first domain for training and the second for evaluation.
We bucket different objects based on the numerical value of their \emph{size} based on their median value in \textsc{DoQ} \cite{elazar-etal-2019-large}, and then manually fix any errors. 
This probe requires prior knowledge of object sizes and understanding of a comparative language construction.
Overall, we collected 127 and 35 objects for training and development respectively. We automatically instantiate object slots using objects that are in the same bucket.

\noindent
\bfemph{Results}
\robertal{} excels in this task, starting from 84\% acc. in the zero-shot setup and reaching \maxmetric{} of 91\% (Table \ref{tab:numbercomparison}). Other models start with random performance and are roughly on par with \mlmbaseline{}. \robertal{} shows sensitivity to the language, suggesting that the ability to compare object sizes is encoded in it.

\noindent
\bfemph{Analysis}
Table \ref{tab:obj-comparison-matrix} shows results of running \robertal{} in the zero-shot setup over pairs of objects, where we sampled a single object from each bucket.
Objects are ordered by their size from small to large. Overall, \robertal{} correctly predicts \nl{larger} below the diagonal, and \nl{smaller} above it. Interestingly, errors are concentrated around the diagonal, due to the more fine-grained differences in sizes, and when we compare objects to \nl{sun}, mostly emitting \nl{larger}, ignoring the rest of the statement.


\begin{table}[h]
\centering
\resizebox{1.0\columnwidth}{!}{
\begin{tabular}{l|c|cc|cc|cc}
 Model & Zero & \multicolumn{2}{c|}{$\MLP{}$} & \multicolumn{2}{c|}{\linear{}}
 & \multicolumn{2}{c}{\langsenses{}}  \\ 
\toprule
& shot &\smetric{}&\maxmetric{}&\smetric{}&\maxmetric{}&\pertlangs{}& \nolangs{}\\ 
\midrule
RoBERTa-L &  84 &  88 &  91 &  86 &  90 &  22 &  26 \\
BERT-WWM  &  55 &  65 &  81 &  63 &  77 &   9 &   9 \\
BERT-L    &  52 &  56 &  66 &  53 &  56 &   5 &   4 \\
\hdashline
BERT-B    &  56 &  55 &  72 &  53 &  56 &   2 &   3 \\
RoBERTa-B &  50 &  61 &  74 &  57 &  66 &   8 &   0 \\
\hdashline
Baseline  &  46 &  57 &  74 &   - &  - &   2 &   1 \\

\end{tabular}}
\caption{Results for the \textsc{Objects Comparison} probe. Accuracy over two answer candidates (random is 50\%).}
\label{tab:numbercomparison}
\end{table}

\begin{table}[h]
\centering
\resizebox{1.0\columnwidth}{!}{
\begin{tabular}{lllllllll}
\toprule
{} &     nail &      pen &   laptop &    table &    house & airplane &     city &      sun \\
\midrule
nail     &                                  - &  \textcolor[HTML]{009F3D}{smaller} &  \textcolor[HTML]{009F3D}{smaller} &  \textcolor[HTML]{009F3D}{smaller} &  \textcolor[HTML]{009F3D}{smaller} &  \textcolor[HTML]{009F3D}{smaller} &  \textcolor[HTML]{009F3D}{smaller} &  \textcolor[HTML]{009F3D}{smaller} \\
pen      &  \textcolor[HTML]{009F3D}{smaller} &                                  - &  \textcolor[HTML]{009F3D}{smaller} &  \textcolor[HTML]{009F3D}{smaller} &  \textcolor[HTML]{009F3D}{smaller} &  \textcolor[HTML]{009F3D}{smaller} &  \textcolor[HTML]{009F3D}{smaller} &  \textcolor[HTML]{009F3D}{smaller} \\
laptop   &   \textcolor[HTML]{DF0024}{larger} &   \textcolor[HTML]{DF0024}{larger} &                                  - &   \textcolor[HTML]{DF0024}{larger} &  \textcolor[HTML]{009F3D}{smaller} &  \textcolor[HTML]{009F3D}{smaller} &  \textcolor[HTML]{009F3D}{smaller} &  \textcolor[HTML]{009F3D}{smaller} \\
table    &   \textcolor[HTML]{DF0024}{larger} &   \textcolor[HTML]{DF0024}{larger} &   \textcolor[HTML]{DF0024}{larger} &                                  - &  \textcolor[HTML]{009F3D}{smaller} &   \textcolor[HTML]{DF0024}{larger} &  \textcolor[HTML]{009F3D}{smaller} &   \textcolor[HTML]{DF0024}{larger} \\
house    &   \textcolor[HTML]{DF0024}{larger} &   \textcolor[HTML]{DF0024}{larger} &   \textcolor[HTML]{DF0024}{larger} &   \textcolor[HTML]{DF0024}{larger} &                                  - &   \textcolor[HTML]{DF0024}{larger} &  \textcolor[HTML]{009F3D}{smaller} &   \textcolor[HTML]{DF0024}{larger} \\
airplane &   \textcolor[HTML]{DF0024}{larger} &   \textcolor[HTML]{DF0024}{larger} &   \textcolor[HTML]{DF0024}{larger} &   \textcolor[HTML]{DF0024}{larger} &   \textcolor[HTML]{DF0024}{larger} &                                  - &   \textcolor[HTML]{DF0024}{larger} &   \textcolor[HTML]{DF0024}{larger} \\
city     &   \textcolor[HTML]{DF0024}{larger} &   \textcolor[HTML]{DF0024}{larger} &   \textcolor[HTML]{DF0024}{larger} &   \textcolor[HTML]{DF0024}{larger} &   \textcolor[HTML]{DF0024}{larger} &   \textcolor[HTML]{DF0024}{larger} &                                  - &   \textcolor[HTML]{DF0024}{larger} \\
sun      &   \textcolor[HTML]{DF0024}{larger} &   \textcolor[HTML]{DF0024}{larger} &   \textcolor[HTML]{DF0024}{larger} &   \textcolor[HTML]{DF0024}{larger} &   \textcolor[HTML]{DF0024}{larger} &   \textcolor[HTML]{DF0024}{larger} &   \textcolor[HTML]{DF0024}{larger} &                                  - \\
\bottomrule
\end{tabular}}
\caption{\robertal{} Zero-shot \textsc{Size comp.} predictions.} 
\label{tab:obj-comparison-matrix}
\end{table}

\subsection{Do LMs know \nl{always} from \nl{often}?}
Adverbial modifiers such as \nl{always}, \nl{sometimes} or \nl{never}, tell us about the quantity or frequency of events~\cite{lewis1975adverbs, barwise1981generalized}.
Anecdotally, when \robertal{} predicts a completion for the phrase \nl{Cats usually drink [MASK].}, the top completion is \nl{coffee}, a frequent drink in the literature it was trained on, rather then \nl{water}.
However, humans know that\nl{Cats NEVER drink coffee}. 
Prior work explored retrieving the correct
quantifier for a statement \cite{herbelot2015building,NIPS2017_6871}. Here we adapt this task to a masked language model.

\noindent
\bfemph{The \textsc{\nl{Always-Never}} task}
We present statements, such as 
\nl{rhinoceros [MASK] have fur}, with answer candidates, such as \nl{never} or \nl{always}. 
To succeed, the model must know the frequency of an event, and map the appropriate adverbial modifier to that representation.
Linguistically, the task tests how well the model predicts frequency quantifiers (or adverbs) modifying predicates in different statements \cite{lepore2007donald}.

\noindent
\bfemph{Probe Construction}
We manually craft templates that contain one slot for a \texttt{subject} and another for an \texttt{object}, e.g. \nl{\texttt{FOOD-TYPE} is \texttt{[MASK]} part of a \texttt{ANIMAL}'s diet.} (more examples available in Table~\ref{tab:model_analysis}). 
The \texttt{subject} slot is instantiated with concepts of the correct semantic type, according to the \texttt{isA} predicate in \textsc{ConceptNet}. In the example above we will find concepts that are of type \texttt{FOOD-TYPE} and \texttt{ANIMAL}. The \texttt{object} slot is then instantiated by forming masked templates of the form \nl{meat is part of a [MASK]'s diet.} and \nl{cats have [MASK].} and letting \bertl{} produce the top-20 completions. We filter out completions that do not have the correct semantic type according to the \texttt{isA} predicate.
Finally, we crowdsource gold answers using Amazon Mechanical Turk. Annotators were presented with an instantiated template (with the masked token removed), such as \nl{Chickens have horns.} and chose the correct answer from $5$ candidates: \nl{never}, \nl{rarely}, \nl{sometimes}, \nl{often} and \nl{always}.\footnote{The class distribution over the answers is \nl{never}:24\%, \nl{rarely}:10\%, \nl{sometimes}:34\%, \nl{often}:7\% and \nl{always}:23\%.}
We collected 1,300 examples with 1,000 used for training and 300 for evaluation. 

We note that some examples in this probe are similar to \textsc{Objects Comparison} (line 4 in Table~\ref{tab:coffeecatsresults}). However, the model must also determine if sizes can be overlapping, which is the case in 56\% of the examples.

\noindent
\bfemph{Results}
Table~\ref{tab:coffeecatsresults} shows the results, where random accuracy is 20\%, and majority vote accuracy is
35.5\%.
In the zero-shot setup, acc. is less than random. In the $\MLP{}$ and \linear{} setup acc. reaches a maximum of 57\% in \bertl{}, but \mlmbaseline{} obtains similar acc., implying that the task was mostly tackled at fine-tuning time, and the pre-trained representations did not contribute much. 
Language controls strengthen this hypothesis, where performance hardly drops in the \pertlanguage{} control 
and slightly drops in the \nolanguage{} control.
Figure~\ref{fig:intro-fig}B compares the learning curve of \robertal{} with controls. 
\mlmbaseline{} consistently outperforms \robertal{}, which display only minor language sensitivity, suggesting that pre-training is not effective for solving this task.

\begin{table}[h]
\centering
\resizebox{1.0\columnwidth}{!}{
\begin{tabular}{l|c|cc|cc|cc}
 Model & Zero & \multicolumn{2}{c|}{$\MLP{}$} & \multicolumn{2}{c|}{\linear{}}
 & \multicolumn{2}{c}{\langsenses{}}  \\ 
\toprule
& shot &\smetric{}&\maxmetric{}&\smetric{}&\maxmetric{}&\pertlangs{}& \nolangs{}\\ 
\midrule
RoBERTa-L &  14 &  44 &  55 &  26 &  41 &   3 &   5 \\
BERT-WWM  &  10 &  46 &  57 &  32 &  52 &   2 &   3 \\
BERT-L    &  22 &  45 &  55 &  36 &  50 &   3 &   8 \\
\hdashline
BERT-B    &  11 &  44 &  56 &  30 &  52 &   3 &   8 \\
RoBERTa-B &  15 &  43 &  53 &  25 &  44 &   2 &   6 \\
\hdashline
Baseline  &  20 &  46 &  56 &  - &  - &   1 &   2 \\
\end{tabular}}
\caption{Results for the \textsc{Always-Never} probe. Accuracy over five answer candidates (random is 20\%).} 
\label{tab:coffeecatsresults}
\end{table}

\paragraph{Analysis}
We generated predictions from the best model, \bertwwm{}, and show analysis results in 
Table~\ref{tab:model_analysis}. For reference, we only selected examples where human majority vote led to the correct answer, and thus the majority vote is near 100\% on these examples.
Although the answers \nl{often} and \nl{rarely} are the gold answer in 19\% of the training data, the LMs predict these answers in less than 1\% of examples.
In the template \nl{A dish with \texttt{FOOD-TYPE} [MASK] contains \texttt{FOOD-TYPE}.} the LM always predicts \nl{sometimes}.
Overall we find models do not perform well. Reporting bias \cite{gordon2013reporting} may play a roll in the inability to correctly determine that \nl{A rhinoceros NEVER has fur.} Interestingly, behavioral research conducted on blind humans shows they exhibit a similar bias \cite{kim2019knowledge}.

\begin{table}[h]
\centering
\resizebox{1.0\columnwidth}{!}{
\begin{tabular}{l|l|l|l}
  \textbf{Question} & \textbf{Answer} & \textbf{Distractor} & \textbf{Acc.} \\ 
\hline
\emph{A dish with \underline{pasta} [MASK] contains \underline{pork} .  } & \textbf{sometimes}     & sometimes   & 75   \\
\hdashline
\emph{\underline{stool} is [MASK] placed in the \underline{box} . }  & never  & \textbf{sometimes}    &  68 \\ 
\hdashline
\emph{A \underline{lizard} [MASK] has a \underline{wing} . }           & never     & \textbf{always}   &  61  \\
\hdashline
\emph{A \underline{pig} is [MASK] smaller than a \underline{cat} .  }   & rarely     & \textbf{always}   & 47   \\
\hdashline
\emph{\underline{meat} is [MASK] part of a \underline{elephant}'s diet .}  & never  & \textbf{sometimes}    & 41  \\
\hdashline
\emph{A \underline{calf} is [MASK] larger than a \underline{dog} .}  & \textbf{sometimes}  &  often    & 30  \\ 
\hdashline
\end{tabular}}
\caption{Error analysis for \textsc{Always-Never}. Model predictions are in bold, and Acc. shows acc. per template.}
\label{tab:model_analysis}
\end{table}


\subsection{Do LMs Capture Negation?}


Ideally, the presence of the word \nl{not} should affect the prediction of a masked token. 
However, Several recent works have shown that LMs
do not take into account the presence of negation in sentences \cite{ettinger2019bert,nie2019adversarial,kassner2019negated}. 
Here, we add to this literature, by probing whether LMs can properly use negation in the context of \emph{synonyms} vs. \emph{antonyms}.

\paragraph{Do LMs Capture the Semantics of Antonyms?}
In the statement  \nl{He was [MASK] fast, he was very slow.}, [MASK] should be replaced with \nl{not}, since \nl{fast} and \nl{slow} are antonyms. 
Conversely, in \nl{He was [MASK] fast, he was very \textbf{rapid}}, the LM should choose a word like \nl{very} in the presence of the synonyms \nl{fast} and \nl{rapid}.       
An LM that correctly distinguishes between \nl{not} and \nl{very}, demonstrates knowledge of the taxonomic relations 
as well as the ability to reason about the usage of negation in this context.

\noindent
\bfemph{Probe Construction}
We sample synonym and antonym pairs from
\textsc{ConceptNet} \cite{speer2017conceptnet} and \textsc{WordNet} \cite{fellbaum1998wordnet}, and use Google Books Corpus to choose pairs that occur frequently in language. We make use of the statements introduced above. Half of the examples are synonym pairs and half antonyms, generating 4,000 training examples and 500 for evaluation.
Linguistically, we test whether the model appropriately predicts a negation vs. intensification adverb based on synonymy/antonymy relations between nouns, adjectives and verbs.

\noindent
\bfemph{Results}
\robertal{} shows higher than chance acc. of 75\% in the zero-shot setting, as well as high \langsense{}
(Table~\ref{tab:negation}). \mlmbaseline{}, equipped with GloVe word embeddings, is able to reach a comparable \smetric{} of 67 and \maxmetric{} of 80\%,
suggesting they do not have a large advantage on this task. 


\begin{table}[h]
\centering
\resizebox{1.0\columnwidth}{!}{
\begin{tabular}{l|c|cc|cc|cc}
 Model & Zero & \multicolumn{2}{c|}{$\MLP{}$} & \multicolumn{2}{c|}{\linear{}}
 & \multicolumn{2}{c}{\langsenses{}}  \\ 
\toprule
& shot &\smetric{}&\maxmetric{}&\smetric{}&\maxmetric{}&\pertlangs{}& \nolangs{}\\  
\midrule
RoBERTa-L &  75 &  85 &  91 &  77 &  84 &  14 &  21 \\
BERT-WWM  &  57 &  70 &  81 &  61 &  73 &   5 &   6 \\
BERT-L    &  51 &  70 &  82 &  58 &  74 &   5 &   9 \\
\hdashline
BERT-B    &  52 &  68 &  81 &  59 &  74 &   2 &   9 \\
RoBERTa-B &  57 &  74 &  87 &  63 &  78 &  10 &  16 \\
\hdashline
Baseline  &  47 &  67 &  80 &   - &  - &   0 &   0 \\
\end{tabular}}
\caption{Results for the \textsc{Antonym  Negation} probe. Accuracy over two answer candidates (random is 50\%).}
\label{tab:negation}
\end{table}

\begin{figure}[t]
  \includegraphics[width=\columnwidth]{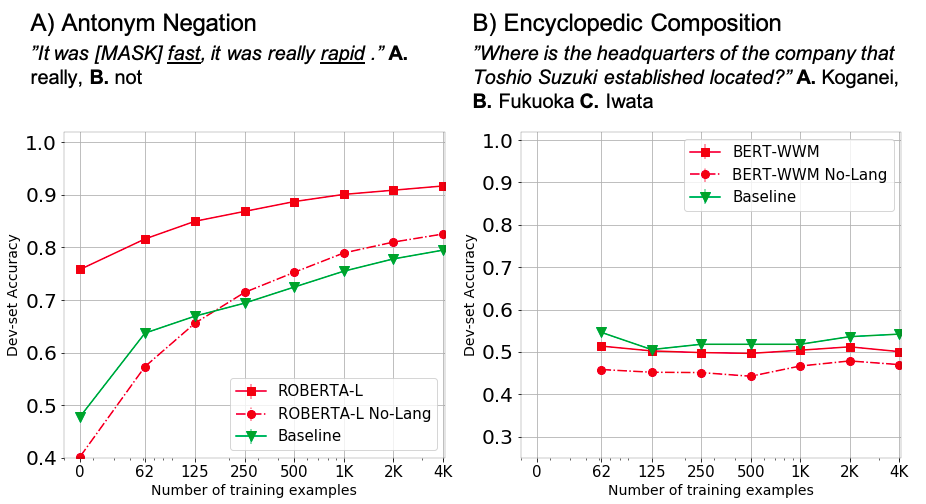}
  \caption{Learning curves in two tasks. For each task the best performing LM is shown alongside the \nolanguage{} control and baseline model. (A) is the correct answer.}~\label{fig:learning-curves}
\end{figure}

\subsection{Can LMs handle conjunctions of facts?}
\label{subsec:conjunction}
We present two probes where a model should understand the reasoning expressed by the word \emph{and}.


\noindent
\paragraph{Property conjunction}
\textsc{ConceptNet} is a Knowledge-Base that describes the properties of millions of concepts through its \texttt{(subject, predicate, object)} triples. We use \textsc{ConcepNet} to test whether LMs can find concepts for which a conjunction of properties holds. For example, we will create a question like \nl{What is located in a street and is related to octagon?}, where the correct answer is \nl{street sign}. Because answers are drawn from \textsc{ConceptNet}, they often consist of more than one word-piece, thus examples are generated in the MC-QA setup.

\noindent
\bfemph{Probe Construction}
To construct an example, we first choose a concept that has two properties in \textsc{ConceptNet}, where a property is a (\texttt{predicate}, \texttt{object}) pair.
For example, \texttt{stop sign} has the properties \texttt{(atLocation,street)} and (\texttt{relatedTo}, \texttt{octagon}). Then, we create two distractor concepts, for which only one property holds: \texttt{car} has the property (\texttt{atLocation}, \texttt{street}), and \texttt{math} has the property (\texttt{relatedTo}, \texttt{octagon}). Given the answer concept, the distractors and the properties, we can automatically generate pseudo-langauge questions and answers by mapping 15 \textsc{ConceptNet} predicates to natural language questions. We split examples such that concepts in training and evaluation are disjoint.
This linguistic structure tests whether the LM can answer questions with conjoined predicates, requiring world knowledge of object and relations.

\noindent
\bfemph{Results}
In MC-QA, we fine-tune the entire network and do not freeze any representations. Zero-shot cannot be applied since the weights of $\MLPQA{}$ are untrained.
All LMs consistely improve as the number of examples increases, reaching a \maxmetric{} of 57-87\% (Table~\ref{tab:property_conjunction}). The high \maxmetric{} results suggest that the LMs generally have the required  pre-existing knowledge.  The \smetric{} of most models is slightly higher than the baselines (49\% \maxmetric{} and 39 \smetric{}). \langsense{} is slightly higher than zero in some models. 
Overall, results suggest the LMs do have some capability in this task, but proximity to baseline results, and low language selectivity make it hard to clearly determine if it existed before fine-tuning.

To further validate our findings, we construct a parallel version of our data, where we replace the word \nl{and} by the phrase \nl{but not}. In this version, the correct answer is the first distractor in the original experiment, where one property holds and the other does not. Overall, we observe a similar trend (with an increase in performance across all models): \maxmetric{} results are high (79-96\%), pointing that the LMs hold the relevant information, but improvement over ESIM-Baseline and language sensitivity are low. For brevity, we omit the detailed numerical results.

\begin{table}[h]
\begin{center}
\resizebox{0.7\columnwidth}{!}{
\begin{tabular}{l|cc|cc}
 Model & \multicolumn{2}{c|}{\textsc{LearnCurve}}  & \multicolumn{2}{c}{\langsenses{}} \\ 
\toprule
&\smetric{}&\maxmetric{}& \pertlangs{}& \nolangs{}  \\ 
\midrule
RoBERTa-L &  49 &  87 &   2 &   4 \\
BERT-WWM  &  46 &  80 &   0 &   1 \\
BERT-L    &  48 &  75 &   2 &   5 \\
\hdashline
BERT-B    &  47 &  71 &   2 &   1 \\
RoBERTa-B &  40 &  57 &   0 &   0 \\
 \hdashline
Baseline  &  39 &  49 &   0 &   0 \\
\end{tabular}
}
\end{center}
\caption{Results for the \textsc{Property Conjunction} probe. Accuracy over three answer candidates (random is 33\%).}
\label{tab:property_conjunction}
\end{table}

\comment{
\begin{table}[h]
\begin{center}
\resizebox{0.7\columnwidth}{!}{
\begin{tabular}{l|cc|cc}
 Model & \multicolumn{2}{c|}{LC}  & \multicolumn{2}{c}{Language} \\ 
\toprule
            &  S  & Max & part& none  \\ 
\midrule
RoBERTa-L &  79 &  93 &   1 &   3 \\
BERT-WWM  &  79 &  94 &   0 &   5 \\
BERT-L    &  71 &  91 &   1 &   1 \\
 \hdashline
BERT-B    &  69 &  92 &   2 &   1 \\
RoBERTa-B &  77 &  92 &   1 &   3 \\
XLNET     &  68 &  88 &   0 &   4 \\
\hdashline
Baseline  &  52 &  78 &   0 &   0 \\
\end{tabular}
}
\end{center}
\caption{Results for the \textsc{Property Conjunction} probe.} 
\label{tab:generalization}
\end{table}
}

\paragraph{Taxonomy conjunction}
A different operation is to find properties that are shared by two concepts. Specifically, we test whether LMs can find the mutual hypernym of a pair of concepts. For example, 
\nl{A germ and a human are both a type of [MASK].}, where the answer is \nl{organism}.

\noindent
\bfemph{Probe Construction}
We use \textsc{ConceptNet} and \textsc{WordNet} to find pairs of concepts and their hypernyms, keeping only pairs that frequently appear in the \textsc{Google Book Corpus}. The example template is \nl{A \texttt{ENT-1} and a \texttt{ENT-2} are both a type of [MASK].}, where \texttt{ENT-1} and \texttt{ENT-2} are replaced with entities that have a common hypernym, which is the gold answer. Distractors are concepts that are hypernyms of \texttt{ENT-1}, but not \texttt{ENT-2}, or vice versa. For evaluation, we keep all examples related to food and animal taxonomies, e.g., \nl{A beer and a ricotta are both a type of [MASK].}, where the answer is \nl{food} and the distractors are \nl{cheese} and \nl{alcohol}.
This phrasing requires the model to handle conjoined co-hyponyms in the subject position, based on lexical relations of hyponymy / hypernymy between nouns.
For training, we use examples from different taxonomic trees, such that the concepts in the training and evaluation sets are disjoint.

\noindent
\bfemph{Results}
Table~\ref{tab:taxonomy} shows that models' zero-shot acc. is substantially higher than random (33\%), but overall even after fine-tuning acc. is at most 59\%. 
However, the \nolanguage{} control shows some language sensitivity, suggesting that some models have pre-existing capabilities. 

\begin{table}[h]
\centering
\resizebox{1.0\columnwidth}{!}{
\begin{tabular}{l|c|cc|cc|cc}
 Model & Zero & \multicolumn{2}{c|}{$\MLP{}$} & \multicolumn{2}{c|}{\linear{}}
 & \multicolumn{2}{c}{\langsenses{}}  \\ 
\toprule
& shot &\smetric{}&\maxmetric{}&\smetric{}&\maxmetric{}&\pertlangs{}& \nolangs{}\\ 
\midrule
RoBERTa-L &  45 &  50 &  56 &  45 &  46 &   0 &   3 \\
BERT-WWM  &  46 &  48 &  52 &  46 &  46 &   0 &   7 \\
BERT-L    &  53 &  54 &  57 &  53 &  54 &   0 &  15 \\
\hdashline
BERT-B    &  47 &  48 &  50 &  47 &  47 &   0 &  12 \\
RoBERTa-B &  46 &  50 &  59 &  47 &  49 &   0 &  18 \\
\hdashline
Baseline  &  33 &  33 &  47 &   - &  - &   1 &   2 \\
\end{tabular}}
\caption{Results for the \textsc{Taxonomy Conjunction} probe. 
Accuracy over three answer candidates (random is 33\%).
} 
\label{tab:taxonomy}
\end{table}

\noindent
\bfemph{Analysis}
Analyzing the errors of \robertal{}, we found that a typical error is predicting for \nl{A crow and a horse are both a type of [MASK].} that the answer is \nl{bird}, rather than \nl{animal}.
Specifically, LMs prefer hypernyms that are closer in terms of edge distance on the taxonomy tree. Thus, a crow is first a bird, and then an animal. We find that when distractors are closer 
to one of the entities in the statement than the gold answer, the models will consistently (80\%) choose the distractor, ignoring the second entity in the phrase. 

\subsection{Can LMs do multi-hop reasoning?}

Questions that require multi-hop reasoning, such as \nl{Who is the director of the movie about a WW2 pacific medic?}, 
have recently drawn attention 
\cite{yang2018hotpotqa,welbl2017constructing,talmor2018web} as a challenging task for contemporary models. 
But do pre-trained LMs have some internal mechanism to handle such questions?

To address this question, we create two probes, one for compositional question answering, and the other uses a multi-hop setup, building upon our observation (\S\ref{sec:controlled_experiments}) that some LMs can compare ages.

\paragraph{Encyclopedic composition}
We construct questions such as \nl{When did the band where John Lennon played first form?}. Here answers require multiple tokens, thus we use the MC-QA setup.


\noindent
\bfemph{Probe Construction}
We use the following three templates: (1) \nl{when did the band where \texttt{ENT} played first form?}, (2) \nl{who is the spouse of the actor that played in \texttt{ENT}?} and (3) \nl{where is the headquarters of the company that \texttt{ENT} established located?}. 
We instantiate \texttt{ENT} using information from \textsc{Wikidata} \cite{vrandecic2014wikidata}, choosing challenging distractors. For example, for template 1, the distractor will be a year close to the gold answer, and for template 3, it will be a city in the same country as the gold answer city.
This linguistic structure introduces a (restrictive) relative clauses that requires a) Correctly resolving the reference of the noun modified by the relative clause, b) Answering the full question subsequently.

To solve the question, the model must have knowledge of all single-hop encyclopedic facts required for answering it. Thus, we first fine-tune the model on all such facts (e.g. \nl{What company did Bill Gates establish? Microsoft}) 
from the training and evaluation set, and then fine-tune on multi-hop composition.

\noindent
\bfemph{Results}
Results are summarized in Table \ref{tab:encyclopedic-composition-res}. All models achieve low acc. in this task, and the baseline performs best with a \maxmetric{} of 54\%. Language sensitivity of all models is small, and \mlmbaseline{} performs slightly better (Figure~\ref{fig:learning-curves}B), suggesting that the LMs are unable to resolve compositional questions, but also struggle to learn it with some supervision.

\begin{table}[h]
\begin{center}
\resizebox{0.7\columnwidth}{!}{
\begin{tabular}{l|cc|cc}
 Model & \multicolumn{2}{c|}{\textsc{LearnCurve}}  & \multicolumn{2}{c}{\langsenses{}} \\ 
\toprule
&\smetric{}&\maxmetric{}& \pertlangs{}& \nolangs{}  \\ 
\midrule
RoBERTa-L &  42 &  50 &   0 &   2 \\
BERT-WWM  &  47 &  53 &   1 &   4 \\
BERT-L    &  45 &  51 &   1 &   4 \\
 \hdashline
BERT-B    &  43 &  48 &   0 &   3 \\
RoBERTa-B &  41 &  46 &   0 &   0 \\
 \hdashline
ESIM-Baseline  &  49 &  54 &   3 &   0 \\

\end{tabular}
}
\end{center}
\caption{Results for \textsc{Encyclopedic composition}. Accuracy over three answer candidates (random is 33\%). 
}
\label{tab:encyclopedic-composition-res}
\end{table}

\paragraph{Multi-hop Comparison}
Multi-hop reasoning can be found in many common structures in natural language. In the phrase \nl{When comparing a 83 year old, a 63 year old and a 56 year old, the [MASK] is oldest}
one must find the oldest person, then refer to its ordering: first, second or third.

\noindent
\bfemph{Probe Construction}
We use the template above, treating the ages as arguments, 
and \nl{first}, \nl{second}, and \nl{third} as answers. Age arguments are in the same ranges as in \agecomp{}.
Linguistically, the task requires predicting the subject of sentences whose predicate is in a superlative form, where the relevant information is contained in a ``when''-clause. The sentence also contains nominal ellipsis, also known as fused-heads \cite{elazar_head}. 

\begin{table}[h]
\centering
\resizebox{1.0\columnwidth}{!}{
\begin{tabular}{l|c|cc|cc|cc}
 Model & Zero & \multicolumn{2}{c|}{$\MLP{}$} & \multicolumn{2}{c|}{\linear{}}
 & \multicolumn{2}{c}{\langsenses{}}  \\ 
\toprule
& shot &\smetric{}&\maxmetric{}&\smetric{}&\maxmetric{}&\pertlangs{}& \nolangs{}\\ 
\midrule
RoBERTa-L &  29 &  36 &  49 &  31 &  41 &   2 &   2 \\
BERT-WWM  &  33 &  41 &  65 &  32 &  36 &   6 &   4 \\
BERT-L    &  33 &  32 &  35 &  31 &  34 &   0 &   3 \\
\hdashline
BERT-B    &  32 &  33 &  35 &  33 &  35 &   0 &   2 \\
RoBERTa-B &  33 &  32 &  40 &  29 &  33 &   0 &   0 \\
\hdashline
Baseline  &  34 &  35 &  48 &   - &  - &   1 &   0 \\
\end{tabular}}
\caption{Results for \textsc{Compositional Comparison}. Accuracy over three answer candidates (random is 33\%). 
}
\label{tab:compositional_comparison}
\end{table}

\noindent
\bfemph{Results}
All three possible answers appear in \robertal{}'s top-10 zero-shot predictions, indicating that the model sees the answers as viable choices. Although successful in  \agecomp{}, the performance of \robertal{} is poor in this probe (Table \ref{tab:compositional_comparison}), With zero-shot acc. that is almost random, \smetric{} slightly above random, \maxmetric{} lower than \mlmbaseline{} (48\%), and close to zero language sensitivity. All LMs seem to be learning the task during probing.
Although \bertwwm{} was able to partially solve the task with a \maxmetric{} of 65\% when approaching 4,000 training examples, the models do not appear to show multi-step capability in this task.

\section{Medals}

\begin{table}[h]
\centering
\resizebox{1.0\columnwidth}{!}{
\begin{tabular}{l|c|c|c|c|c}
& RoBERTa & BERT & BERT & RoBERTa & BERT  \\
& Large & WWM & Large & Base & Base  \\
\hline
\textsc{Always-Never} &  &  & &  &  \\
\hdashline
\textsc{Age Comparison} & \checkmark & \checkmark &  & \semicheck &  \\
\textsc{Objects Compar.} & \checkmark & \semicheck & &  & \\ 
\hdashline
\textsc{Antonym Neg.} & \checkmark & & \semicheck & \semicheck&  \\
\hdashline
\textsc{Property Conj.} & \semicheck & \semicheck &  &  &  \\
\textsc{Taxonomy Conj.} & \semicheck & \semicheck &  & \semicheck &  \\
\hdashline
\textsc{Encyc. Comp.} &  &  &  &  &  \\
\textsc{Multi-hop Comp.} &  &  &  &  & 
\end{tabular}}
\caption{The oLMpic games medals', summarizing per-task success. \checkmark{} indicate the LM has achieved high accuracy considering controls and baselines, \semicheck{} indicates partial success.} 
\label{tab:medals}
\end{table}

We summarize the results of the oLMpic Games in Table \ref{tab:medals}.
Generally, the LMs did not demonstrate strong pre-training capabilities in these symbolic reasoning tasks.
\bertwwm{} showed partial success in a few tasks, whereas
\robertal{} showed high performance in \textsc{Always-Never}, \textsc{Objects Comparison} and \textsc{Antonym Negation}, and emerges as the most promising LM. However, when perturbed, \robertal{} has failed to demonstrates consistent generalization and abstraction.

\paragraph{Analysis of correlation with pre-training data}
A possible hypothesis for why a particular model is successful in a particular task might be that the language of a probe is more common in the corpus it was pre-trained on. To check that, we compute the unigram distribution over the training corpus of both \bert{} and \roberta{}. We then compute the average log probability of the development set under these two unigram distributions for each task (taking into account only content words). Finally, we compute the correlation between which model performs better on a probe (\robertal{} vs. \bertwwm{}) and which training corpus induces higher average log probability on that probe. We find that the Spearman correlation is 0.22, hinting that the unigram distributions do not fully explain the difference in performance.

\comment{
\begin{table*}[t]
\centering
\resizebox{0.8\textwidth}{!}{
\begin{tabular}{l|c|c|c|c|c}
& BERT-B & BERT-L & BERT-WWM & RoBERTa-B & RoBERTa-L \\ 
\hline
Always-Never &  &  & &  &  \\
\hdashline
Numeric Comparison &  &  & \semicheck &  & \checkmark \\
Objects Comparison &  &  & \semicheck &  & \checkmark \\ 
\hdashline
Antonym-Synonym Negation & & \semicheck &  & \semicheck & \checkmark \\
\hdashline
Property Conjunction &  & \semicheck &  &  &  \\
Taxonomy Conjunction &  &  &  &  &  \\
\hdashline
Encyclopedic Composition &  &  &  &  &  \\
Multi-hop Comparison &  &  &  &  & 
\end{tabular}}
\caption{oLMpics summary. Outlining the success / failure of each model on every game we tested. \checkmark{} indicates the model knows the specific game we tested on and \semicheck{} indicates the model manage partially on it.} 
\label{tab:medals}
\end{table*}
}

\section{Discussion}
We presented eight different tasks for evaluating the reasoning abilities of models, alongside an evaluation protocol for disentangling pre-training from fine-tuning. We found that even models that have identical structure and objective functions differ not only quantitatively but also qualitatively. Specifically, \robertal{} has shown reasoning abilities that are absent from other models. Thus, with appropriate data and optimization, models can acquire from an LM objective skills that might be surprising intuitively.

However, when current LMs succeed in a reasoning task, they do not do so through abstraction and composition as humans perceive it. The abilities are context-dependent, if ages are compared -- then the numbers should be typical ages. Discrepancies from the training distribution lead to large drops in performance. Last, the performance of LM in many reasoning tasks is poor.

Our work sheds light on some of the blind spots of current LMs. We will release our code and data to help researchers evaluate the reasoning abilities of models, aid the design of new probes, and guide future work on pre-training, objective functions and model design for endowing models with capabilities they are currently lacking.
\paragraph{Acknowledgements}
This work was completed in partial fulfillment for the PhD degree of the first author. 
We thank our colleagues at The Allen Institute of AI, especially Kyle Richardson, Asaf Amrami, Mor Pipek, Myle Ott, Hillel Taub-Tabib and Reut Tsarfaty.
This research was partially supported by The Israel Science Foundation grant 942/16, The Blavatnik Computer Science Research Fund and The Yandex Initiative for Machine Learning, and the European Union’s Seventh Framework Programme (FP7) under grant agreement no. 802774-ERC-iEXTRACT and no. 802800-DELPHI.

\bibliography{tacl2018}

\begin{thebibliography}{58}
\expandafter\ifx\csname natexlab\endcsname\relax\def\natexlab#1{#1}\fi

\bibitem[{Adi et~al.(2016)Adi, Kermany, Belinkov, Lavi, and
  Goldberg}]{adi2016fine}
Yossi Adi, Einat Kermany, Yonatan Belinkov, Ofer Lavi, and Yoav Goldberg. 2016.
\newblock Fine-grained analysis of sentence embeddings using auxiliary
  prediction tasks.
\newblock \emph{arXiv preprint arXiv:1608.04207}.

\bibitem[{Bagherinezhad et~al.(2016)Bagherinezhad, Hajishirzi, Choi, and
  Farhadi}]{bagherinezhad2016elephants}
Hessam Bagherinezhad, Hannaneh Hajishirzi, Yejin Choi, and Ali Farhadi. 2016.
\newblock Are elephants bigger than butterflies? reasoning about sizes of
  objects.
\newblock In \emph{Thirtieth AAAI Conference on Artificial Intelligence}.

\bibitem[{Barwise and Cooper(1981)}]{barwise1981generalized}
Jon Barwise and Robin Cooper. 1981.
\newblock Generalized quantifiers and natural language.
\newblock In \emph{Philosophy, language, and artificial intelligence}, pages
  241--301. Springer.

\bibitem[{Belinkov and Glass(2019)}]{belinkov2019analysis}
Yonatan Belinkov and James Glass. 2019.
\newblock Analysis methods in neural language processing: A survey.
\newblock \emph{Transactions of the Association for Computational Linguistics},
  7:49--72.

\bibitem[{Blier and Ollivier(2018)}]{blier2018description}
L{\'e}onard Blier and Yann Ollivier. 2018.
\newblock The description length of deep learning models.
\newblock In \emph{Advances in Neural Information Processing Systems}, pages
  2216--2226.

\bibitem[{Chen et~al.(2017)Chen, Zhu, Ling, Wei, Jiang, and
  Inkpen}]{chen2017enhanced}
Qian Chen, Xiaodan Zhu, Zhen-Hua Ling, Si~Wei, Hui Jiang, and Diana Inkpen.
  2017.
\newblock Enhanced {LSTM} for natural language inference.
\newblock In \emph{Proceedings of the 55th Annual Meeting of the Association
  for Computational Linguistics (Volume 1: Long Papers)}, pages 1657--1668,
  Vancouver, Canada. Association for Computational Linguistics.

\bibitem[{Coenen et~al.(2019)Coenen, Reif, Yuan, Kim, Pearce, Vi{\'e}gas, and
  Wattenberg}]{coenen2019visualizing}
Andy Coenen, Emily Reif, Ann Yuan, Been Kim, Adam Pearce, Fernanda Vi{\'e}gas,
  and Martin Wattenberg. 2019.
\newblock Visualizing and measuring the geometry of bert.
\newblock \emph{arXiv preprint arXiv:1906.02715}.

\bibitem[{Dai and Le(2015)}]{NIPS2015_5949}
Andrew~M Dai and Quoc~V Le. 2015.
\newblock \href
  {http://papers.nips.cc/paper/5949-semi-supervised-sequence-learning.pdf}
  {Semi-supervised sequence learning}.
\newblock In C.~Cortes, N.~D. Lawrence, D.~D. Lee, M.~Sugiyama, and R.~Garnett,
  editors, \emph{Advances in Neural Information Processing Systems 28}, pages
  3079--3087. Curran Associates, Inc.

\bibitem[{Devlin et~al.(2019)Devlin, Chang, Lee, and
  Toutanova}]{devlin2019bert}
J.~Devlin, M.~Chang, K.~Lee, and K.~Toutanova. 2019.
\newblock Bert: Pre-training of deep bidirectional transformers for language
  understanding.
\newblock In \emph{North American Association for Computational Linguistics
  (NAACL)}.

\bibitem[{Elazar and Goldberg(2019)}]{elazar_head}
Yanai Elazar and Yoav Goldberg. 2019.
\newblock \href {https://doi.org/10.1162/tacl\_a\_00280} {Where’s my head?
  definition, data set, and models for numeric fused-head identification and
  resolution}.
\newblock \emph{Transactions of the Association for Computational Linguistics},
  7:519--535.

\bibitem[{Elazar et~al.(2019)Elazar, Mahabal, Ramachandran, Bedrax-Weiss, and
  Roth}]{elazar-etal-2019-large}
Yanai Elazar, Abhijit Mahabal, Deepak Ramachandran, Tania Bedrax-Weiss, and Dan
  Roth. 2019.
\newblock \href {https://doi.org/10.18653/v1/P19-1388} {How large are lions?
  inducing distributions over quantitative attributes}.
\newblock In \emph{Proceedings of the 57th Annual Meeting of the Association
  for Computational Linguistics}, pages 3973--3983, Florence, Italy.
  Association for Computational Linguistics.

\bibitem[{Ettinger(2019)}]{ettinger2019bert}
Allyson Ettinger. 2019.
\newblock What bert is not: Lessons from a new suite of psycholinguistic
  diagnostics for language models.
\newblock \emph{arXiv preprint arXiv:1907.13528}.

\bibitem[{Ettinger et~al.(2016)Ettinger, Elgohary, and
  Resnik}]{ettinger2016probing}
Allyson Ettinger, Ahmed Elgohary, and Philip Resnik. 2016.
\newblock Probing for semantic evidence of composition by means of simple
  classification tasks.
\newblock In \emph{Proceedings of the 1st Workshop on Evaluating Vector-Space
  Representations for NLP}, pages 134--139.

\bibitem[{Fellbaum(1998)}]{fellbaum1998wordnet}
C.~Fellbaum. 1998.
\newblock \emph{WordNet: An Electronic Lexical Database}.
\newblock MIT Press.

\bibitem[{Forbes and Choi(2017)}]{forbes2017verb}
Maxwell Forbes and Yejin Choi. 2017.
\newblock Verb physics: Relative physical knowledge of actions and objects.
\newblock In \emph{Proceedings of the 55th Annual Meeting of the Association
  for Computational Linguistics (Volume 1: Long Papers)}, pages 266--276.

\bibitem[{Gardner et~al.(2019{\natexlab{a}})Gardner, Berant, Hajishirzi,
  Talmor, and Min}]{gardner2019making}
Matt Gardner, Jonathan Berant, Hannaneh Hajishirzi, Alon Talmor, and Sewon Min.
  2019{\natexlab{a}}.
\newblock On making reading comprehension more comprehensive.
\newblock In \emph{Proceedings of the 2nd Workshop on Machine Reading for
  Question Answering}, pages 105--112.

\bibitem[{Gardner et~al.(2019{\natexlab{b}})Gardner, Berant, Hajishirzi,
  Talmor, and Min}]{gardner2019question}
Matt Gardner, Jonathan Berant, Hannaneh Hajishirzi, Alon Talmor, and Sewon Min.
  2019{\natexlab{b}}.
\newblock Question answering is a format; when is it useful?
\newblock \emph{arXiv preprint arXiv:1909.11291}.

\bibitem[{Goldberg(2019)}]{goldberg2019assessing}
Yoav Goldberg. 2019.
\newblock Assessing bert's syntactic abilities.
\newblock \emph{arXiv preprint arXiv:1901.05287}.

\bibitem[{Gordon and Van~Durme(2013)}]{gordon2013reporting}
Jonathan Gordon and Benjamin Van~Durme. 2013.
\newblock Reporting bias and knowledge acquisition.
\newblock In \emph{Proceedings of the 2013 workshop on Automated knowledge base
  construction}, pages 25--30. ACM.

\bibitem[{Herbelot and Vecchi(2015)}]{herbelot2015building}
Aur{\'e}lie Herbelot and Eva~Maria Vecchi. 2015.
\newblock Building a shared world: Mapping distributional to model-theoretic
  semantic spaces.
\newblock In \emph{Proceedings of the 2015 Conference on Empirical Methods in
  Natural Language Processing}, pages 22--32.

\bibitem[{Hewitt and Liang(2019)}]{hewitt2019designing}
John Hewitt and Percy Liang. 2019.
\newblock Designing and interpreting probes with control tasks.
\newblock In \emph{Proceedings of the 2019 Conference on Empirical Methods in
  Natural Language Processing and the 9th International Joint Conference on
  Natural Language Processing (EMNLP-IJCNLP)}, pages 2733--2743.

\bibitem[{Hewitt and Manning(2019)}]{structural-probe}
John Hewitt and Christopher~D. Manning. 2019.
\newblock A structural probe for finding syntax in word representations.
\newblock In \emph{Proceedings of the Conference of the North American Chapter
  of the Association for Computational Linguistics: Human Language
  Technologies, {NAACL-HLT}}, pages 4129--4138.

\bibitem[{Jiang et~al.(2019)Jiang, Xu, Araki, and Neubig}]{jiangHowCanWe2019}
Zhengbao Jiang, Frank~F. Xu, Jun Araki, and Graham Neubig. 2019.
\newblock How can we know what language models know?
\newblock \emph{arXiv preprint arXiv:1911.12543}.

\bibitem[{Kassner and Sch{\"u}tze(2020)}]{kassner2019negated}
Nora Kassner and Hinrich Sch{\"u}tze. 2020.
\newblock \href {https://www.aclweb.org/anthology/2020.acl-main.698} {Negated
  and misprimed probes for pretrained language models: Birds can talk, but
  cannot fly}.
\newblock In \emph{Proceedings of the 58th Annual Meeting of the Association
  for Computational Linguistics}, pages 7811--7818, Online. Association for
  Computational Linguistics.

\bibitem[{Kim et~al.(2019)Kim, Elli, and Bedny}]{kim2019knowledge}
Judy~S Kim, Giulia~V Elli, and Marina Bedny. 2019.
\newblock Knowledge of animal appearance among sighted and blind adults.
\newblock \emph{Proceedings of the National Academy of Sciences},
  116(23):11213--11222.

\bibitem[{Lepore and Ludwig(2007)}]{lepore2007donald}
Ernest Lepore and Kirk Ludwig. 2007.
\newblock \emph{Donald Davidson's truth-theoretic semantics}.
\newblock Oxford University Press.

\bibitem[{Lewis(1975)}]{lewis1975adverbs}
David Lewis. 1975.
\newblock Adverbs of quantification.
\newblock \emph{Formal semantics-the essential readings}, 178:188.

\bibitem[{Lin et~al.(2019)Lin, Tan, and Frank}]{lin2019open}
Yongjie Lin, Yi~Chern Tan, and Robert Frank. 2019.
\newblock Open sesame: Getting inside bert’s linguistic knowledge.
\newblock In \emph{Proceedings of the 2019 ACL Workshop BlackboxNLP: Analyzing
  and Interpreting Neural Networks for NLP}, pages 241--253.

\bibitem[{Linzen et~al.(2016{\natexlab{a}})Linzen, Dupoux, and
  Goldberg}]{linzen2016agreement}
Tal Linzen, Emmanuel Dupoux, and Yoav Goldberg. 2016{\natexlab{a}}.
\newblock Assessing the ability of {LSTM}s to learn syntax-sensitive
  dependencies.
\newblock \emph{{TACL}}, 4:521--535.

\bibitem[{Linzen et~al.(2016{\natexlab{b}})Linzen, Emmanuel, and
  Yoav}]{linzen2016assessing}
Tal Linzen, D.~Emmanuel, and G.~Yoav. 2016{\natexlab{b}}.
\newblock Assessing the ability of {LSTMs} to learn syntax-sensitive
  dependencies.
\newblock \emph{Transactions of the Association for Computational Linguistics
  (TACL)}, 4.

\bibitem[{Liu et~al.(2019)Liu, Ott, Goyal, Du, Joshi, Chen, Levy, Lewis,
  Zettlemoyer, and Stoyanov}]{liu2019roberta}
Yinhan Liu, Myle Ott, Naman Goyal, Jingfei Du, Mandar Joshi, Danqi Chen, Omer
  Levy, Mike Lewis, Luke Zettlemoyer, and Veselin Stoyanov. 2019.
\newblock Roberta: A robustly optimized bert pretraining approach.
\newblock \emph{arXiv preprint arXiv:1907.11692}.

\bibitem[{Mihaylov et~al.(2018)Mihaylov, Clark, Khot, and
  Sabharwal}]{OpenBookQA2018}
Todor Mihaylov, Peter Clark, Tushar Khot, and Ashish Sabharwal. 2018.
\newblock Can a suit of armor conduct electricity? a new dataset for open book
  question answering.
\newblock In \emph{EMNLP}.

\bibitem[{Nie et~al.(2020)Nie, Williams, Dinan, Bansal, Weston, and
  Kiela}]{nie2019adversarial}
Yixin Nie, Adina Williams, Emily Dinan, Mohit Bansal, Jason Weston, and Douwe
  Kiela. 2020.
\newblock \href {https://www.aclweb.org/anthology/2020.acl-main.441}
  {Adversarial {NLI}: A new benchmark for natural language understanding}.
\newblock In \emph{Proceedings of the 58th Annual Meeting of the Association
  for Computational Linguistics}, pages 4885--4901, Online. Association for
  Computational Linguistics.

\bibitem[{Pennington et~al.(2014)Pennington, Socher, and
  Manning}]{pennington2014glove}
J.~Pennington, R.~Socher, and C.~D. Manning. 2014.
\newblock Glo{V}e: Global vectors for word representation.
\newblock In \emph{Empirical Methods in Natural Language Processing (EMNLP)},
  pages 1532--1543.

\bibitem[{Peters et~al.(2018{\natexlab{a}})Peters, Neumann, Iyyer, Gardner,
  Clark, Lee, and Zettlemoyer}]{peters2018elmo}
M.~E. Peters, M.~Neumann, M.~Iyyer, M.~Gardner, C.~Clark, K.~Lee, and
  L.~Zettlemoyer. 2018{\natexlab{a}}.
\newblock Deep contextualized word representations.
\newblock In \emph{North American Association for Computational Linguistics
  (NAACL)}.

\bibitem[{Peters et~al.(2018{\natexlab{b}})Peters, Neumann, Zettlemoyer, and
  Yih}]{peters2018dissecting}
Matthew Peters, Mark Neumann, Luke Zettlemoyer, and Wen-tau Yih.
  2018{\natexlab{b}}.
\newblock Dissecting contextual word embeddings: Architecture and
  representation.
\newblock In \emph{Proceedings of the 2018 Conference on Empirical Methods in
  Natural Language Processing}, pages 1499--1509.

\bibitem[{Petroni et~al.(2019)Petroni, Rockt{\"a}schel, Riedel, Lewis, Bakhtin,
  Wu, and Miller}]{petroni2019language}
Fabio Petroni, Tim Rockt{\"a}schel, Sebastian Riedel, Patrick Lewis, Anton
  Bakhtin, Yuxiang Wu, and Alexander Miller. 2019.
\newblock Language models as knowledge bases?
\newblock In \emph{Proceedings of the 2019 Conference on Empirical Methods in
  Natural Language Processing and the 9th International Joint Conference on
  Natural Language Processing (EMNLP-IJCNLP)}, pages 2463--2473.

\bibitem[{Pezzelle and Fern{\'a}ndez(2019)}]{pezzelle2019red}
Sandro Pezzelle and Raquel Fern{\'a}ndez. 2019.
\newblock Is the red square big? malevic: Modeling adjectives leveraging visual
  contexts.
\newblock In \emph{Proceedings of the 2019 Conference on Empirical Methods in
  Natural Language Processing and the 9th International Joint Conference on
  Natural Language Processing (EMNLP-IJCNLP)}, pages 2858--2869.

\bibitem[{Radford et~al.(2019)Radford, Wu, Child, Luan, Amodei, and
  Sutskever}]{radford2019gpt2}
Alec Radford, Jeffrey Wu, Rewon Child, David Luan, Dario Amodei, and Ilya
  Sutskever. 2019.
\newblock Language models are unsupervised multitask learners.
\newblock \emph{OpenAI Blog}, 1(8).

\bibitem[{Rozen et~al.(2019)Rozen, Shwartz, Aharoni, and
  Dagan}]{rozen2019diversify}
Ohad Rozen, Vered Shwartz, Roee Aharoni, and Ido Dagan. 2019.
\newblock Diversify your datasets: Analyzing generalization via controlled
  variance in adversarial datasets.
\newblock In \emph{Proceedings of the 23rd Conference on Computational Natural
  Language Learning (CoNLL)}, pages 196--205.

\bibitem[{Sennrich et~al.(2016)Sennrich, Haddow, and
  Birch}]{sennrich2015neural}
Rico Sennrich, Barry Haddow, and Alexandra Birch. 2016.
\newblock Neural machine translation of rare words with subword units.
\newblock In \emph{Proceedings of the 54th Annual Meeting of the Association
  for Computational Linguistics (Volume 1: Long Papers)}, pages 1715--1725.

\bibitem[{Shwartz and Dagan(2019)}]{lexcomp_tacl_2019}
Vered Shwartz and Ido Dagan. 2019.
\newblock Still a pain in the neck: Evaluating text representations on lexical
  composition.
\newblock In \emph{Transactions of the Association for Computational
  Linguistics (TACL)}.

\bibitem[{Speer et~al.(2017)Speer, Chin, and Havasi}]{speer2017conceptnet}
Robyn Speer, Joshua Chin, and Catherine Havasi. 2017.
\newblock Conceptnet 5.5: An open multilingual graph of general knowledge.
\newblock In \emph{Thirty-First AAAI Conference on Artificial Intelligence}.

\bibitem[{Talmor and Berant(2018)}]{talmor2018web}
A.~Talmor and J.~Berant. 2018.
\newblock The web as knowledge-base for answering complex questions.
\newblock In \emph{North American Association for Computational Linguistics
  (NAACL)}.

\bibitem[{Talmor et~al.(2019)Talmor, Herzig, Lourie, and
  Berant}]{talmor2019commonsenseqa}
A.~Talmor, J.~Herzig, N.~Lourie, and J.~Berant. 2019.
\newblock Commonsenseqa: A question answering challenge targeting commonsense
  knowledge.
\newblock In \emph{North American Association for Computational Linguistics
  (NAACL)}.

\bibitem[{Tenney et~al.(2019{\natexlab{a}})Tenney, Das, and
  Pavlick}]{tenney-etal-2019-bert}
Ian Tenney, Dipanjan Das, and Ellie Pavlick. 2019{\natexlab{a}}.
\newblock \href {https://doi.org/10.18653/v1/P19-1452} {{BERT} rediscovers the
  classical {NLP} pipeline}.
\newblock In \emph{Proceedings of the 57th Annual Meeting of the Association
  for Computational Linguistics}, pages 4593--4601, Florence, Italy.
  Association for Computational Linguistics.

\bibitem[{Tenney et~al.(2019{\natexlab{b}})Tenney, Xia, Chen, Wang, Poliak,
  McCoy, Kim, Durme, Bowman, Das, and Pavlick}]{tenney2018what}
Ian Tenney, Patrick Xia, Berlin Chen, Alex Wang, Adam Poliak, R~Thomas McCoy,
  Najoung Kim, Benjamin~Van Durme, Sam Bowman, Dipanjan Das, and Ellie Pavlick.
  2019{\natexlab{b}}.
\newblock \href {https://openreview.net/forum?id=SJzSgnRcKX} {What do you learn
  from context? probing for sentence structure in contextualized word
  representations}.
\newblock In \emph{International Conference on Learning Representations}.

\bibitem[{Vaswani et~al.(2017)Vaswani, Shazeer, Parmar, Uszkoreit, Jones,
  Gomez, Kaiser, and Polosukhin}]{vaswani2017attention}
Ashish Vaswani, Noam Shazeer, Niki Parmar, Jakob Uszkoreit, Llion Jones,
  Aidan~N Gomez, {\L}ukasz Kaiser, and Illia Polosukhin. 2017.
\newblock Attention is all you need.
\newblock In \emph{Advances in neural information processing systems}, pages
  5998--6008.

\bibitem[{Vrande\v{c}i\'{c} and Kr\H{o}tzsch(2014)}]{vrandecic2014wikidata}
D.~Vrande\v{c}i\'{c} and M.~Kr\H{o}tzsch. 2014.
\newblock Wikidata: A free collaborative knowledgebase.
\newblock \emph{Communications of the ACM}, 57.

\bibitem[{Wallace et~al.(2019)Wallace, Wang, Li, Singh, and
  Gardner}]{wallace2019nlp}
Eric Wallace, Yizhong Wang, Sujian Li, Sameer Singh, and Matt Gardner. 2019.
\newblock Do nlp models know numbers? probing numeracy in embeddings.
\newblock In \emph{Proceedings of the 2019 Conference on Empirical Methods in
  Natural Language Processing and the 9th International Joint Conference on
  Natural Language Processing (EMNLP-IJCNLP)}, pages 5310--5318.

\bibitem[{Wang et~al.(2017)Wang, Tang, Wang, and Deng}]{NIPS2017_6871}
Mingzhe Wang, Yihe Tang, Jian Wang, and Jia Deng. 2017.
\newblock \href
  {http://papers.nips.cc/paper/6871-premise-selection-for-theorem-proving-by-deep-graph-embedding.pdf}
  {Premise selection for theorem proving by deep graph embedding}.
\newblock In I.~Guyon, U.~V. Luxburg, S.~Bengio, H.~Wallach, R.~Fergus,
  S.~Vishwanathan, and R.~Garnett, editors, \emph{Advances in Neural
  Information Processing Systems 30}, pages 2786--2796. Curran Associates, Inc.

\bibitem[{Warstadt et~al.(2019)Warstadt, Cao, Grosu, Peng, Blix, Nie, Alsop,
  Bordia, Liu, Parrish et~al.}]{warstadt2019investigating}
Alex Warstadt, Yu~Cao, Ioana Grosu, Wei Peng, Hagen Blix, Yining Nie, Anna
  Alsop, Shikha Bordia, Haokun Liu, Alicia Parrish, et~al. 2019.
\newblock Investigating bert’s knowledge of language: Five analysis methods
  with npis.
\newblock In \emph{Proceedings of the 2019 Conference on Empirical Methods in
  Natural Language Processing and the 9th International Joint Conference on
  Natural Language Processing (EMNLP-IJCNLP)}, pages 2870--2880.

\bibitem[{Welbl et~al.(2018)Welbl, Stenetorp, and
  Riedel}]{welbl2017constructing}
Johannes Welbl, Pontus Stenetorp, and Sebastian Riedel. 2018.
\newblock Constructing datasets for multi-hop reading comprehension across
  documents.
\newblock \emph{Transactions of the Association for Computational Linguistics},
  6:287--302.

\bibitem[{Yang et~al.(2018{\natexlab{a}})Yang, Birnbaum, Wang, and
  Downey}]{yang2018extracting}
Yiben Yang, Larry Birnbaum, Ji-Ping Wang, and Doug Downey. 2018{\natexlab{a}}.
\newblock Extracting commonsense properties from embeddings with limited human
  guidance.
\newblock In \emph{Proceedings of the 56th Annual Meeting of the Association
  for Computational Linguistics (Volume 2: Short Papers)}, pages 644--649.

\bibitem[{Yang et~al.(2018{\natexlab{b}})Yang, Qi, Zhang, Bengio, Cohen,
  Salakhutdinov, and Manning}]{yang2018hotpotqa}
Z.~Yang, P.~Qi, S.~Zhang, Y.~Bengio, W.~W. Cohen, R.~Salakhutdinov, and C.~D.
  Manning. 2018{\natexlab{b}}.
\newblock {HotpotQA}: A dataset for diverse, explainable multi-hop question
  answering.
\newblock In \emph{Empirical Methods in Natural Language Processing (EMNLP)}.

\bibitem[{Yang et~al.(2019)Yang, Dai, Yang, Carbonell, Salakhutdinov, and
  Le}]{yang2019xlnet}
Zhilin Yang, Zihang Dai, Yiming Yang, Jaime Carbonell, Russ~R Salakhutdinov,
  and Quoc~V Le. 2019.
\newblock Xlnet: Generalized autoregressive pretraining for language
  understanding.
\newblock In \emph{Advances in neural information processing systems}, pages
  5753--5763.

\bibitem[{Yogatama et~al.(2019)Yogatama, de~M.~d'Autume, Connor, Kocisky,
  Chrzanowski, Kong, Lazaridou, Ling, Yu, Dyer et~al.}]{yogatama2019learning}
D.~Yogatama, C.~de~M.~d'Autume, J.~Connor, T.~Kocisky, M.~Chrzanowski, L.~Kong,
  A.~Lazaridou, W.~Ling, L.~Yu, C.~Dyer, et~al. 2019.
\newblock Learning and evaluating general linguistic intelligence.
\newblock \emph{arXiv preprint arXiv:1901.11373}.

\bibitem[{Zellers et~al.(2018)Zellers, Bisk, Schwartz, and
  Choi}]{zellers2018swag}
Rowan Zellers, Yonatan Bisk, Roy Schwartz, and Yejin Choi. 2018.
\newblock Swag: A large-scale adversarial dataset for grounded commonsense
  inference.
\newblock In \emph{Proceedings of the 2018 Conference on Empirical Methods in
  Natural Language Processing (EMNLP)}.

\end{thebibliography}
\bibliographystyle{acl_natbib}

\end{document}